# Recognition Confidence Analysis of Handwritten Chinese Character with CNN

Meijun He, Shuye Zhang, Huiyun Mao, Lianwen Jin
School of Electronic and Information Engineering
South China University of Technology
Guangzhou, China
wenjum@gmail.com, shuye.chueng@gmail.com, huiyun.mao01@gmail.com, eelwjin@scut.edu.cn

*Abstract*—In this paper, we present an effective method to analyze the recognition confidence of handwritten Chinese character, based on the softmax regression score of a high performance convolutional neural networks (CNN). Through careful and thorough statistics of 827,685 testing samples that randomly selected from total 8836 different classes of Chinese characters, we find that the confidence measurement based on CNN is an useful metric to know how reliable the recognition results are. Furthermore, we find by experiments that the recognition confidence can be used to find out similar and confusable character-pairs, to check wrongly or cursively written samples, and even to discover and correct mis-labelled samples. Many interesting observations and statistics are given and analyzed in this study.

## I. INTRODUCTION

Handwritten Chinese character recognition has received intensive attention for decades and makes great progress in the past 40 years [1-5, 10-15]. Some traditional methods including 8-directional features [2], and modified quadratic discriminant function (MQDF) [3] [4] has demonstrated their effectiveness. Recently, the deep convolutional neural networks [5] and some effective techniques which aim to prevent overfitting, such as dropout [6], are widely used and show promising performance in the fields of computer vision and pattern recognition, such as image classification and detection [7]. At the same time, some researchers apply CNNs to Chinese handwriting character recognition field and also achieve exciting performance [8] [9].

Deep convolutional neural networks need millions of labeled character samples during training process [7]. With the rapid growth of handwriting character samples from Internet, it makes training a deep model possible. However, the labeled data from Internet is not very reliable and the human annotation is too expensive. A feasible way to solve the problem is the automatic labeling based on some criteria such as the recognition confidence. Confidence transformation was used for classifier combination by Lin et al [10] and Liu et al [11]. Wang [12] applied confidence transformation for improving handwritten Chinese text recognition. However, we still don't know how to make use of the confidence and how reliable it is in the situation of handwritten data mining and ground-truth labeling. Thus, efficient and effective methods that can analyze and mine character data are urgently required.

In this paper, we present an effective method to analyze the recognition confidence of handwritten Chinese character, based on the softmax regression score of a high performance convolutional neural networks (CNN). This method can be used for automatically discovering reliable labeled-data. Our experimental results show that we can utilize the recognition confidence to analyze a large amount of character samples and find many useful knowledge.

## II. CLASSIFIER DESIGN AND CONFIDENCE MEASUREMENT.

We have designed a deep convolutional neural network (CNN) for large-scale handwritten Chinese character recognition (HCCR) following a similar architecture of the DeepCNet proposed by Graham [16]. The architecture of our network is represented as follows:

$96 \times 96 Input\text{-}100C3\text{-}MP2\text{-}200C2\text{-}MP2\text{-}300C2\text{-}MP2\text{-}400C2\text{-}MP2\text{-}500C2\text{-}MP2\text{-}600C2\text{-}1024FC\text{-}10081Output$.

An online handwritten character is rendered into a $96 \times 96$ bitmap and then inputted to the first convolutional layer, which contains 100 convolutional filter kernels of size $3 \times 3$. The following pooling layer takes 100 previous feature maps as input, and applies 100 max-pooling kernels of size $2 \times 2$ (denoted as MP2). The second convolutional layer has 200 feature maps. Each one is obtained by filtering the feature maps in the first pooling layer with 200 kernels of size $2 \times 2$, then followed by a max pooling layer of MP2. The third convolutional layer has 300 convolutional kernels, followed by a MP2, and so on. After the final convolutional layer (600C2), one full connection layer with 1024 neuros is applied, following by a softmax output layer.

We adopt ReLU non-linearity as activation function between convolutional layer and pooling layer [17]. Meanwhile, dropout, as an important and effective method proposed by [18], is used to prevent overfitting and improve recognition accuracy.

Our CNN classifier can recognize as many as 10,081 classes that contain mixtures of handwritten Chinese, English letters, digits, and symbols. It is worth noting that some pre-processing techniques, such as the path signature feature extraction [16] [19], data argumentation with deformation transformation [20], were used to train the CNN classifier. Give an inputing handwritten character, each-class confidence reflects the reliability that one classifier recognizes it as the corresponding class. We employ convolutional neural network softmax regression as the output layer, and the softmax regression output is defined as a confidence metric in the paper. The output confidence is given by

$$h_\theta(x^{(i)}) = \begin{bmatrix} p(y^{(i)}=1|x^{(i)};\theta) \\ p(y^{(i)}=2|x^{(i)};\theta) \\ \vdots \\ p(y^{(i)}=k|x^{(i)};\theta) \end{bmatrix} = \frac{1}{\sum_{j=1}^{k} e^{\theta_j^T x^{(i)}}} \begin{bmatrix} e^{\theta_1^T x^{(i)}} \\ e^{\theta_2^T x^{(i)}} \\ \vdots \\ e^{\theta_k^T x^{(i)}} \end{bmatrix} \quad (1)$$

where $p(y^{(i)} = j | x^{(i)};\theta)$ is the $i$ character's confidence for the $j$ output category. $\theta_j^T$ denotes the weights connected to the $j$ output category in last fully connected layer. It is worth noting that the confidence metric given by the CNN takes value between [0,1], which can also be regarded as a kind of confidence probability.

## III. DESCRIPTION OF TESTING DATASETS.

We used approximately 10 million samples for training the CNN classifier and 2 million samples for validation, which were primarily selected from the SCUT-COUCH dataset [15], the CASIA OLHWDB1.0-1.2 [14] dataset, and some in-house datasets. It took approximately four weeks for us to train and optimize the deep CNN. Testing was conducted on another 827,685 samples, which were randomly selected from seven datasets as shown in Table 1. The seven testing datasets are 863 [13], OLHWDB-CASIA1.0 [14], OLHWDB-CASIA1.1 [14], OLHWDB-CASIA1.2 [14], SCUT-COUCH [15], HKU [2] dataset and a unpublished dataset named "In-house"

The testing dataset contains 827,685 character samples of 8836 different classes, includes all simplified Chinese characters, traditional Chinese characters, and some rarely-used Chinese characters. The sample number of any character class varies from 3 to 163.

TABLE I.   TEST DATASETS USED IN OUR EXPERIMENTS

| Name | #Samples | #Character Categories |
|---|---|---|
| 863 | 40578 | 6763 |
| OLHWDB-CASIA1.0 | 143600 | 3865 |
| OLHWDB-CASIA1.1 | 98235 | 3755 |
| OLHWDB-CASIA1.2 | 79154 | 2923 |
| SCUT-COUCH | 161166 | 8836 |
| HKU | 184089 | 6763 |
| In-house | 120863 | 6763 |
| Total | 827685 | 8836 |

## IV. EXPERIMENTAL RESULTS AND ANALYSIS

The experimental results will be illustrated and analyzed in three parts. In part A, the recognition rate of each testing dataset will be presented. The character categories whose recognition rate is highest or lowest are extracted out and analyzed accordingly. In part B, a statistic is applied to approximate the confidence distribution and expose the relationship between confidence and recognition rate. In part C, recognition confidence information is utilized to mine and discover some interesting knowledge from the testing data, such as samples mistakenly labeled, samples written cursively, and similar and confusable characters sets.

### A. The Statics of the Recognition Accuracy

The recognition rate on the testing datasets is given in TABLE II. It can be seen that we achieve a very exciting average recognition rates of 97.73%, 99.12% and 99.88% for top 1, top 2 and top 10 recognition candidates, respectively. This shows the great potential and classification ability of the CNN based classifier. We made some statistics on the recognition accuracy distribution against different classes of characters, and we found that there are 2935 classes whose recognition rate reach 100% and 1077 classes among them contain more 50 characters samples for each. For illustration, 20 character categories randomly extracted from the 1077 categories are given in TABLE III, along with the corresponding examples of samples shown in Fig. 1. Similarly, 20 characters with the lowest accuracy are shown in TABLE IV and their corresponding samples are given in Fig. 2.

From TABLE III and Fig. 1, we found that those character categories with high recognition rate mostly have such characteristics as follows: their confidence is usually higher than 0.98, the stroke-structure is well-organized, and there are seldom confusable character against them. Thus even though some samples are cursively written, the classifier can still correctly recognize them with high confidence.

On the contrast, for the characters with low accuracy, we found they have some characteristics in common, which are summarized below:

- Some characters are very confusable with symbols. For instance, characters such as "丶", "丨", "一", "入" and "丫" are easily confused by symbols "、", "1", "—", "λ" and "Y", and vice versa.

- The small difference of stroke length or stroke direction form some similar character pairs make them very difficult to be distinguished. For example, characters "囗", "汩", "攵", "巳", "毁", "沫" and "睢" are easily to be recognized as corresponding similar character of "口", "汨", "夂", "己", "毁", "沬" and "睢", respectively.

- The tiny distinction of stroke structure bring challenges for some similar character pairs, such as "佘 - 佘", "睛 - 晴" and "海 - 诲", owing to the structure "入", "目" and "氵" is very similar and confusable to "人", "日" and "讠" respectively.

The above phenomena may be the main reason why characters of such kinds of classes are very hard to be recognized. Besides, we also note that the average confidence of these classes is commonly lower, and the stroke number is usually much less than those character classes with high recognition rates.

TABLE II. RECOGNITION RATE (%).

| Top Candidates | 1 | 2 | 3 | 5 | 10 |
|---|---|---|---|---|---|
| 863 | 99.72 | 99.94 | 99.99 | 100.00 | 100.00 |
| OLHWDB-CASIA1.0 | 96.69 | 98.60 | 99.20 | 99.55 | 99.77 |
| OLHWDB-CASIA1.1 | 96.04 | 98.35 | 99.04 | 99.47 | 99.75 |
| OLHWDB-CASIA1.2 | 97.24 | 98.86 | 99.34 | 99.66 | 99.84 |
| SCUT-COUCH | 98.62 | 99.58 | 99.80 | 99.92 | 99.96 |
| HKU | 97.70 | 99.40 | 99.68 | 99.84 | 99.93 |
| In-house | 98.41 | 99.19 | 99.56 | 99.76 | 99.89 |
| Average | 97.73 | 99.12 | 99.51 | 99.74 | 99.88 |

TABLE III. 20 CHARACTER CATEGORIES THAT RECOGNITION RATE IS 100%.

| Character Category | Average Confidence | Average Stroke Number |
|---|---|---|
| 僵 | 0.9996 | 11.9 |
| 瓢 | 0.9992 | 14.4 |
| 剔 | 0.9988 | 10.1 |
| 藤 | 0.9983 | 12.8 |
| 蚤 | 0.9983 | 6.5 |
| 南 | 0.9980 | 6.3 |
| 楔 | 0.9977 | 10.7 |
| 裔 | 0.9967 | 9.0 |
| 帆 | 0.9961 | 4.9 |
| 券 | 0.9949 | 6.5 |
| 豹 | 0.9948 | 7.8 |
| 匙 | 0.9944 | 7.5 |
| 粥 | 0.9943 | 7.0 |
| 艾 | 0.9941 | 4.6 |
| 暴 | 0.9932 | 9.8 |
| 碧 | 0.9929 | 8.2 |
| 鼠 | 0.9920 | 10.6 |
| 瓶 | 0.9910 | 7.8 |
| 梨 | 0.9906 | 8.4 |
| 兜 | 0.9889 | 8.3 |

TABLE IV. 20 CHARACTER CATEGORIES THAT RECOGNITION RATE IS LOWEST.

| Character Category | Recognition Rate(%) | Average Confidence | Average Stroke Number |
|---|---|---|---|
| 丨 | 23.46 | 0.4346 | 1.2 |
| 一 | 66.25 | 0.4525 | 1.0 |
| 丶 | 23.08 | 0.4743 | 1.2 |
| 呐 | 61.07 | 0.6507 | 5.0 |
| 泪 | 50.55 | 0.6525 | 6.5 |
| 口 | 41.61 | 0.6604 | 2.2 |
| 子 | 76.92 | 0.6903 | 1.7 |
| 入 | 66.67 | 0.7080 | 2.0 |
| 已 | 69.47 | 0.7251 | 2.4 |
| 内 | 76.16 | 0.7371 | 3.6 |
| 丫 | 72.90 | 0.7473 | 2.3 |
| 女 | 69.32 | 0.7658 | 3.2 |
| 毁 | 74.83 | 0.7799 | 9.4 |
| 丁 | 77.07 | 0.7827 | 1.7 |
| 沫 | 75.76 | 0.7854 | 7.0 |
| 雎 | 75.61 | 0.7920 | 10.8 |
| 睛 | 74.15 | 0.8092 | 9.5 |
| 海 | 77.18 | 0.8440 | 6.9 |
| 计 | 74.29 | 0.8697 | 3.5 |
| 汆 | 66.67 | 0.8769 | 5.0 |

TABLE V. THE SAMPLE DISTRIBUTION ON CONFIDENCE AND THE ACCURACY DISTRIBUTION CONFIDENCE.

| Confidence | Samples (%) | Accuracy (%) |
|---|---|---|
| [0.0,0.1) | 0.00 | 0.00 |
| [0.1,0.2) | 0.01 | 13.73 |
| [0.2,0.3) | 0.09 | 20.46 |
| [0.3,0.4) | 0.24 | 33.00 |
| [0.4,0.5) | 0.48 | 42.79 |
| [0.5,0.6) | 1.16 | 53.56 |
| [0.6,0.7) | 1.28 | 64.68 |
| [0.7,0.8) | 1.62 | 74.82 |
| [0.8,0.9) | 2.65 | 85.11 |
| [0.9,0.95) | 2.89 | 92.51 |
| [0.95,0.97) | 2.38 | 95.75 |
| [0.97,0.98) | 2.09 | 97.31 |
| [0.98,0.99) | 3.96 | 98.27 |
| [0.99,0.999) | 16.65 | 99.40 |
| [0.999,0.9999) | 19.35 | 99.88 |
| [0.9999,1] | 45.15 | 99.97 |

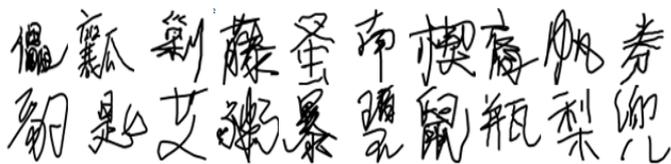

Fig. 1. Character categories whose recognition rate is 100%.

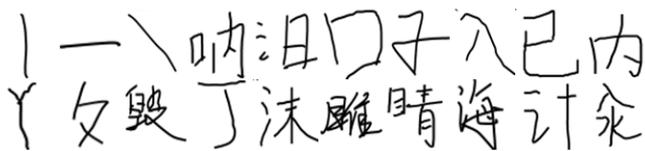

Fig. 2. Character categories with lowest recognition rate.

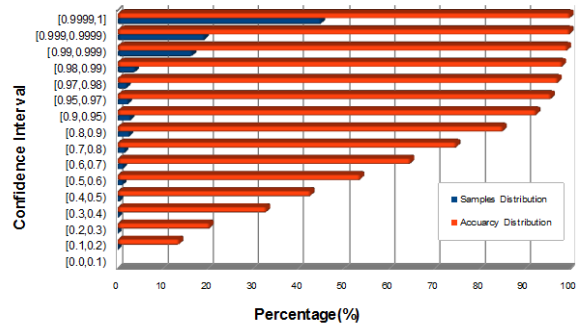

Fig. 3. Samples distribution and Accuracy on Confidence Interval chart.

## B. Recognition Confidence Statistics

The sample distribution on confidence and the accuracy distribution on confidence are given on TABLE V. As the confidence of most characters is distributed in high confidence interval, we apply dense quantization in high confidence interval. Figure 3 shows the distribution of recognition accuracy against confidence interval, with corresponding samples distribution.

From Table V, several interesting observations can be concluded. First, we can see that samples are mainly distributed in confidence range of [0.9, 1], which contains as high as 92.46% of the testing samples. Particularly, there are 45% of testing samples with very high recognition accuracy locate in the confidence interval of [0.9999, 1]. Second, the higher confidence comes along with the higher accuracy. When the confidence given by classifier is higher than 0.99, the accuracy can reach up to 99.75% in average, which means that if the CNN gives recognition result with confidence higher than 0.99, we have a high probability of 99.75% to tell that the recognition result is correct. Besides, a testing sample whose confidence is higher than 0.9999 may be correctly recognized with 99.97% probability (in fact, we can see in subsequent section that the 0.03% mis-recognized characters are all mis-labelled samples). On the contrast, a testing sample with less than 0.5 confidence has less than 22% possibility to be recognized correctly. It indicates that we may take focus on those samples with low confidence for improving classifier recognition performance.

## C. The Utility of Recognition Confidence

Through careful observation and analysis of the recognition confidence for handwritten Chinese character based on CNN, we found certain kinds of samples/knowledge can be mined from the testing datasets. In Part B, it is shown that the higher confidence comes along with the higher accuracy. So we analyze the samples whose confidence is higher than 0.9999 but incorrectly recognized, which are shown in Fig. 4.

It can be seen from Fig.4 that, in fact, these samples are labeled wrongly and the CNN classifier actually give the correct recognition results. With the help of the confidence by this way, we find out those incorrectly labeled samples in the testing dataset. The situation of samples whose recognition confidence is higher than 0.99 but incorrectly recognized is presented in Table VI.

Fig. 4. Character samples are presented in the first row. The corresponding original label is in the second row and the predicting result by our deep convolutional neural networks is in the third row.

TABLE VI. DESCRIPTIONS ABOUT SAMPLES WITH CONFIDENCE HIGHER THAN 0.999 BUT RECOGNIZED ERROR

| Confidence Interval | [0.99,0.999) | [0.999,0.9999) | [0.9999,1] |
|---|---|---|---|
| #Samples | 137791 | 160136 | 373658 |
| # incorrectly recognized before correction | 824 | 186 | 97 |
| #Labeled incorrectly annotated | 434 | 177 | 97 |
| Accuracy after corrected | 99.68% | 99.99% | 100.00% |

It can be seen that the original labels of most of them are in fact incorrect. After correcting mislabeled characters, the final accuracy by the recognizer achieve 99.99% and 100% for characters with confidence ranges [0.999, 0.9999] and [0.9999, 1] respectively. Moreover, 45.15% samples are located on the confidence interval [0.9999, 1] with almost 100% recognition rate. Therefore, the CNN recognizer can be used to help to automatically label those samples whose confidence is in interval [0.9999, 1]. It's of great significance for automatically annotating large scale of character samples from Internet.

We randomly investigate some samples whose confidence is lower than 0.2, and some examples are illustrated in Fig. 5. From Fig.5, it can be summarized with several phenomena that may lead to low confidence:

- Heavy cursively handwritten styles such as sample 2, 6, 7, 10, 13, 14, 15.
- Low resolution such as sample 3 and 4.
- Wrongly written such as sample 8,11,12,17.
- Skew character such as sample 16.

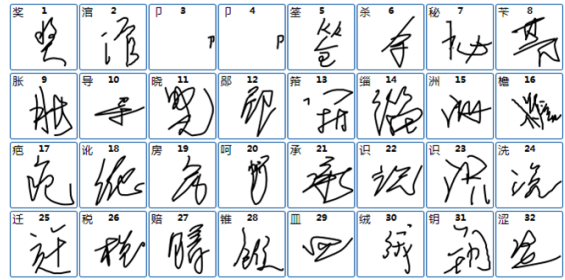

Fig. 5. Samples whose confidence is lower than 0.2

By observing these incorrectly recognized samples, it gives us some clues on what kinds of samples are not easily handled by CNN classifier, and the CNN classifier needs what kinds of of training data.

We can also make use of the recognition confidence to find similar Chinese characters. The diagram for seeking similar character is shown in Fig. 6. For a given character category, we firstly collect character samples whose top 1 candidates are the same. Second, the confidence of the top 10 recognition candidates are accumulated. Third, the candidate categories are sorted in descending order. Finally, we select the second candidate as the similar candidate.

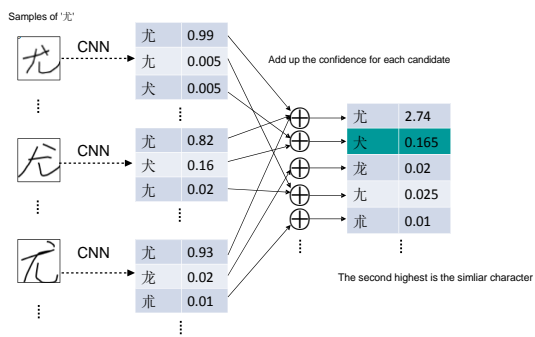

Fig. 6. The process of seeking similar characters.

Fig. 7 shows some similar character pairs generated by this method and their corresponding samples are shown in Fig.8, which indicates that this method for finding similar character pairs is reasonable and effective.

**Similar Character Pairs**

犬 — 尤　汩 — 汩　戌 — 戌　秃 — 秃
壬 — 王　呐 — 呐　讠 — 氵　棟 — 楝
日 — 曰　毀 — 毀　胄 — 胄　廿 — 卄
子 — 孑　丢 — 丢　斤 — 厅　忄 — 小
口 — 口　千 — 干　诋 — 泜　疗 — 疗

Fig. 7. Similar characters pairs.

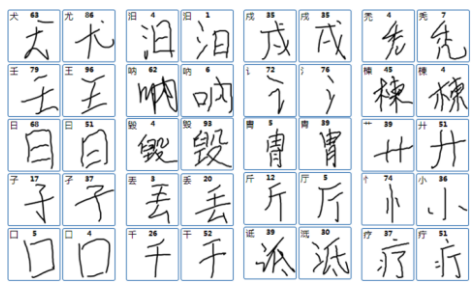

Fig. 8. Samples of similar characters pairs.

## V. CONCLUSIONS

In this paper, we present a thorough analysis of the recognition confidence of handwritten Chinese characters based on convolutional neural networks. Through this study, we can draw the following conclusions:

1. The confidence measurement based on CNN is effective approach for us to know how reliable the recognition results are. Usually, high confidence means high probability of the CNN to generate correct recognition results. In particular, we found that if the confidence is larger than 0.9999, we can say with almost 100% certainty that the recognition result is correct.

2. For handwritten Chinese character recognition, more than 92% characters can be recognized with high confidence. This shows that the CNN classifier is very effective and reliable.

3. The recognition confidence can be used to find and correct mis-labeled, wrongly or cursively written samples automatically.

4. The recognition confidence can be used to find similar and confusable character-pairs.